\documentclass[sigconf]{acmart}

\AtBeginDocument{%
  }

\usepackage{microtype}
\usepackage{graphicx}
\usepackage{subfigure}
\usepackage{tabularx}
\usepackage{booktabs} 
\usepackage{multirow}
\usepackage{soul}
\usepackage{array}
\usepackage{hyperref}
\usepackage{amsmath}
\usepackage{mathtools}
\usepackage{amsthm}

\copyrightyear{2023}
\acmYear{2023}
\setcopyright{acmlicensed}
\acmConference[CIKM '23]{Proceedings of the 32nd ACM International Conference on Information and Knowledge Management}{October 21--25, 2023}{Birmingham, United Kingdom}
\acmBooktitle{Proceedings of the 32nd ACM International Conference on Information and Knowledge Management (CIKM '23), October 21--25, 2023, Birmingham, United Kingdom}
\acmPrice{15.00}
\acmDOI{10.1145/3583780.3615193}
\acmISBN{979-8-4007-0124-5/23/10}

\begin{document}

\title{Effective Slogan Generation with Noise Perturbation}


\author{Jongeun Kim}
\affiliation{%
 \institution{UNIST}
 \city{Ulsan}
 \country{Republic of Korea}}
\email{joannekim5456@unist.ac.kr}

\author{MinChung Kim}
\affiliation{%
 \institution{UNIST}
 \city{Ulsan}
 \country{Republic of Korea}}
\email{mckim@unist.ac.kr}

\author{Taehwan Kim}
\affiliation{%
 \institution{UNIST}
 \city{Ulsan}
 \country{Republic of Korea}}
\email{taehwankim@unist.ac.kr}

\renewcommand{\shortauthors}{Jongeun Kim, MinChung Kim, \& Taehwan Kim} 

\begin{abstract}
   Slogans play a crucial role in building the brand's identity of the firm. A slogan is expected to reflect firm's vision and the brand's value propositions in memorable and likeable ways. Automating the generation of slogans with such characteristics is challenging. Previous studies developed and tested \textit{slogan generation} with syntactic control and summarization models which are not capable of generating distinctive slogans. We introduce a novel approach that leverages pre-trained transformer T5 model with noise perturbation on newly proposed 1:N matching pair dataset. This approach serves as a contributing factor in generating distinctive and coherent slogans. Furthermore, the proposed approach incorporates descriptions about the firm and brand into the generation of slogans. We evaluate generated slogans based on ROUGE-1, ROUGE-L and Cosine Similarity metrics and also assess them with human subjects in terms of slogan's distinctiveness, coherence, and fluency. The results demonstrate that our approach yields better performance than baseline models and other transformer-based models. 
\end{abstract}




\maketitle

\section{Introduction} \label{sec1}
Slogan is a key element of a brand’s identity \cite{keller1993conceptualizing}. A good slogan is supposed to reflect the firm’s vision and the brand’s value proposition in memorable and likeable way \cite{kohli2013you, dass2014study}. Traditionally, to generate a slogan, human experts go through the process of brainstorming, generating and evaluating candidate slogans, requiring a lot of resources and manpower. Automation in generating candidate slogans can alleviate a significant amount of labor and resources  \cite{misawa2020distinctive, jin2023towards, alnajjar2021computational, tomavsic2014implementation}. Recognizing the importance of automatic slogan generation, prior researchers compiled different datasets and developed algorithms applicable to the datasets. In particular, \cite{ozbal2013brainsup, tomavsic2014implementation, alnajjar2021computational} explored slogan generation in syntactic perspective; they focused on filling in a well-informed sentence skeleton extracted from a slogan database with suitable words. Another line of research utilized slogan descriptions \cite{chandler2018slogatron, znidarvsic2015case}, while recently, \cite{misawa2020distinctive, jin2023towards, ccougalmics2022generating} adopted seq2seq(sequence-to-sequence) models for automatic slogan generation, using descriptions of firms as inputs.

Several prior studies highlight the importance of creativity and diversity as key factors for creating memorable and likeable slogan. For instance, \cite{jin2023towards} proposed generating syntactically diverse slogan while maintaining the truthfulness to the source text. To achieve these two characteristics, the method finetunes an abstractive summarization model initialized on CNN/DailyMail dataset \cite{hermann2015teaching} with syntactic control. However, although it may preserve the truthfulness to source description, abstractive summarization model tends to adhere to the abstractive features, which does not generate diverse and creative slogans. The approach presented in \cite{misawa2020distinctive} addresses this problem by defining and minimizing a reconstruction loss as an indicator of distinctiveness. By leveraging a seq2seq model on a Japanese job description-slogan paired dataset with the reconstruction loss, this method pursues to generate unique and specific slogans to the given description. Even so, this approach heavily  relies on the size of a dataset when a PLM(pre-trained language model) is not available, and lacks domain coverage for diverse firm level slogans.

In order to alleviate these issues, our focus lies in adopting more general PLM combined with noise perturbation applied to input embeddings. Noise perturbation has been extensively studied in various domains such as robustness against adversarial attacks, data augmentation, improvement of generation quality, and learning sentence representations \cite{wu2022frsum, ravishankar2022effects, savinov2021step, wu2021smoothed, zhou2022sharpness, zhu2019freelb, zhang2018word, aghajanyan2020better}. Building upon the insights gained from these approaches, we introduce Gaussian noise perturbation to continuous input embeddings as to facilitate diverse and coherent generation. We further conduct an ablation study to determine the most successful amount of noise perturbation. Additionally, we compare the results from our approach with prior works with ROUGE-1, ROUGE-L scores and Cosine Similarity as evaluation metrics that compute the correlations between descriptions and reference slogans with generated slogans.

In summary, our main contributions are as follows:
\vspace{-0.3cm}
\begin{itemize}
    \item We propose and further release a new refined dataset in a broader domain, which may promote subsequent work.
    \item We incorporate noise perturbation into transformer-based model to enhance the coherence, fluency, and distinctiveness of generated slogans.
    \item We provide a comprehensive evaluation of different models, encompassing evaluation metrics, human evaluations and qualitative analysis. Our approach performs best in quantitative and qualitative results.
\end{itemize}
\section{Method}
\subsection{Dataset}
Following the previous studies \cite{jin2023towards, misawa2020distinctive, ccougalmics2022generating}, our transformer-based model uses the same data format which includes descriptions about the firm or target brands as input and slogan as output. The authors of \cite{jin2023towards} complied such descriptions and slogans by web crawling based on a dataset comprising company details(i.e. domain, industry, and linkedin url). However, the dataset provided contained numerous irrelevant firm descriptions, incorrect slogans, and noises due to the crawling of HTML <meta> tags. Moreover, the dataset used in \cite{misawa2020distinctive} is not an open-source and is narrowed only to the job matching domain, limiting its applicability in broader contexts. 

Thus, in order to create a comprehensive dataset consisting of accurate and uncontaminated description-slogan pairs, we collect our data from a well-known slogan database website\footnote{https://www.sloganlist.com/}. We crawl descriptions and slogans with 16 different categories (e.g. food, drinks, tours, cosmetics, technology, etc). To enable diverse generation, we employ 1:N matching, where a single firm description is paired with multiple brand slogans. Subsequently, we pre-process the dataset obtained from crawling to remove unnecessary special characters and firm foundation dates. Our final dataset comprise a total of 12K pairs, out of which 1.2K are used as the test set.

\subsection{Baselines}
We adopt \cite{misawa2020distinctive, jin2023towards} as our baselines, representing non-PLM and summarization PLM respectively. The first baseline \cite{misawa2020distinctive} introduces a reconstruction model based on compressed representation to generate slogans with distinctiveness. Additionally, it adds refer–attention and conversion layers to account for topic emphasis and style difference. However, since this work targets generating slogans constrained to job matching domain, we train their method using our dataset. Following the original settings, we train with randomly initialized embeddings, where noise perturbation is unnecessary and expand the definition of distinctiveness consisting the concept of creativeness.

For our second baseline \cite{jin2023towards}, we use the same pre-trained abstractive summarization model, which freezes the encoder layer up to the second last layer, including the input embedding layers. When applying noise perturbation approach to the second baseline model, we modify the model by freezing the encoder layers but updating the gradients in the input embeddings.
\subsection{Noise Perturbation}
\begin{figure}[ht]
  \centering

    \includegraphics[width=0.8\columnwidth]{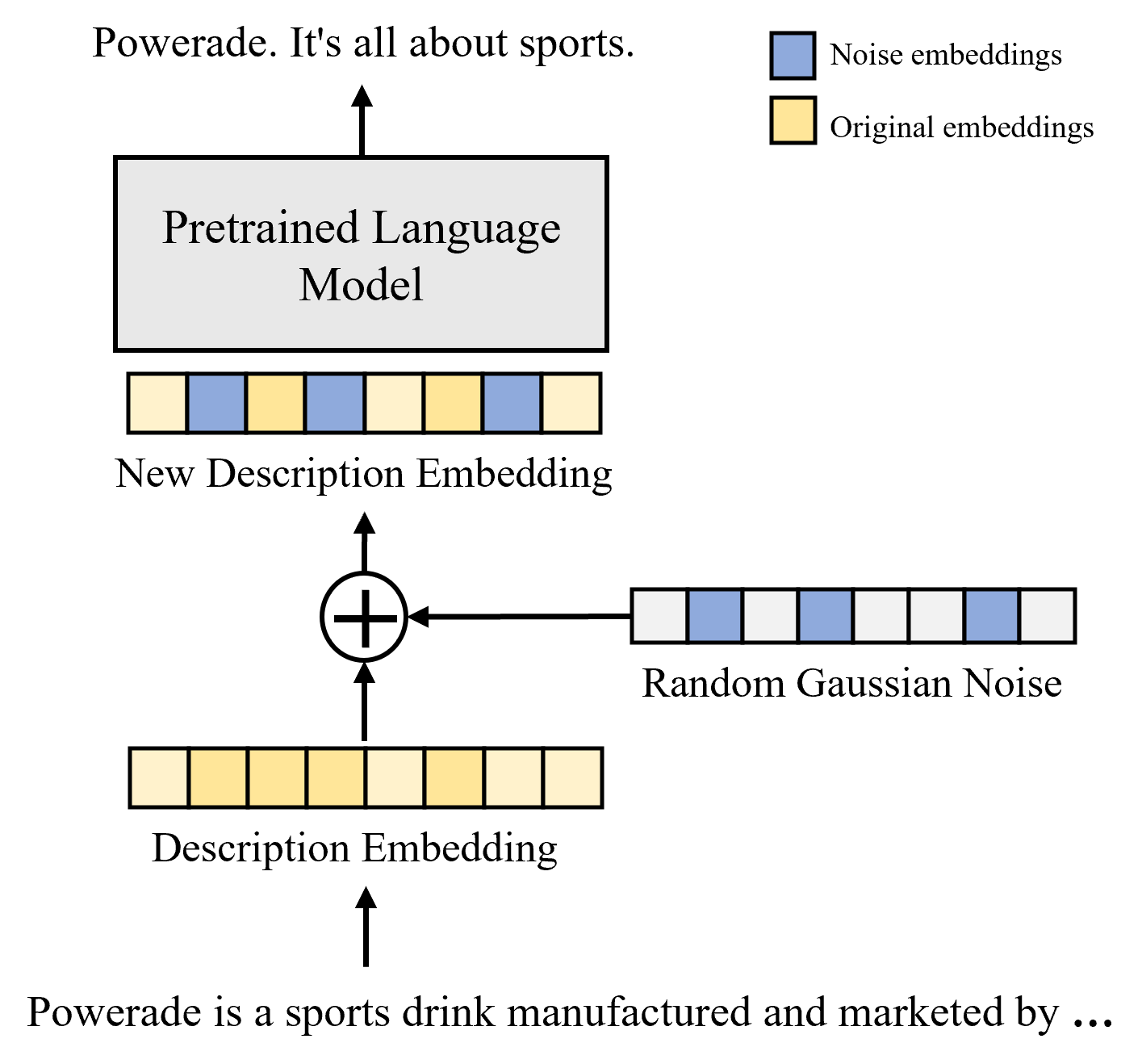}
    \caption{Model architecture of random Gaussian noise perturbed to description input embeddings before being fed into PLM.}
  \label{model architecture}
  \vspace{-0.5em}
\end{figure}
To apply noise perturbation on PLMs, we select three transformer-based models: BART \cite{lewis2019bart}, GPT2 \cite{radford2019language}, and T5 \cite{raffel2020exploring}. To align with the approach presented in \cite{jin2023towards}, which trains on distil-BART, we incorporate our noise perturbation method using the proposed settings for the purpose of baseline comparison. GPT2, which is a decoder-only architecture, is particularly employed to experiment as a variation of the encoder-decoder architecture. Lastly, we leverage the text-to-text T5 transformer, which we expect to be better suited for our description-slogan (text-to-text) paired dataset. This selection aligns with our implementation, where the perturbed input embeddings can fully undergo comprehensive processing through the encoder before being channeled into the decoder.

Drawing inspiration from \cite{savinov2021step, zhu2019freelb, zhang2018word, zhou2022sharpness}, with specific emphasis on the $\delta$-SAM \cite{zhou2022sharpness} strategy which empirically demonstrates that perturbing embedding yields better results compared to perturbing the loss function, we implement noise perturbation into the continuous embedding space of input embeddings prior to feeding it into the PLM. This is in contrast to perturbing the discrete words themselves. While previous studies have utilized noise for data augmentation, enhancing representation learning, and improving generation quality, to the best of our knowledge, we are the first to posit that noise perturbation can effectively stimulate diverse generation. This proposition is grounded in its likely shared relationship with representation learning and data augmentation. Our empirical support for this conjecture is provided in Section \ref{exper}.

\begin{equation}
    \label{eq1}
     X_{noise} \sim \mathcal{N}(I, \sigma^2I)
\end{equation}
\begin{equation}
    \label{eq2}
    X_{emb} = \begin{cases}
     &  X_{noise_i} \text{\,\,, \,\,\,\,if} \,\, i \,\,\, \text{in} \,\,\ Pr_{\theta_{noise}}\\ 
     &  X_{emb_i} \text{\,\,, \,\,\,\,\,else}\,\,\,\,\,\,\,\,\,  
    \end{cases} \\
\end{equation}

To provide details on our noise perturbation technique in Figure \ref{model architecture}, we utilize a hyperparameter $\theta_{noise}$ to determine the level of noise, which we further explain in \ref{absection}. Specifically, following normal random Gaussian noise with mean 0 and shared variance $\sigma^2I$ in Equation\ref{eq1}, we randomly select number of $i$ tokens from total $X_{emb}$ tokens with the probability of $\theta_{noise}$. Next, the selected tokens are  substituted with these Gaussian noise as $X_{noise}$ in Equation\ref{eq2}, before being passed to the PLM.

\begin{table*} \centering
  \caption{Average scores from five experiments of the method and baselines on the test dataset. The higher the better for each column, best scores indicated in bold. Desc and Ref each denotes to firm/brand descriptions and reference slogans.}
  \vspace{-0.3cm}
  \label{mainperf}
  \begin{tabular}{p{3.5cm}|w{c}{1.5cm}|w{c}{1.5cm}|w{c}{1.5cm}|w{c}{1.5cm}|w{c}{1.5cm}|w{c}{1.5cm}}\toprule 
    \multirow{2}{*}{Model} &\multicolumn{2}{|c}{ROUGE-1} &\multicolumn{2}{|c}{ROUGE-L} &\multicolumn{2}{|c}{Cosine Similarity}\\
\cline{2-7}
     & Desc & Ref & Desc & Ref & Desc & Ref \\
    \midrule
    Reconstruction \cite{misawa2020distinctive} & 6.2680 & 5.8606 & 5.5827 & 5.6405 & 8.2791 & 17.3551 \\ [0.5ex]
    \cline{1-7} 
    BART \cite{jin2023towards} & 16.1577 & 8.8890 & 13.6226 & 8.3501 & 41.0308 & 29.5885 \\
    BART + noise & 15.6148 & 8.1342 & 13.1228 & 7.4970 & 40.9499 & 28.3322 \\[0.5ex]
    \cline{1-7} 
    GPT2  & 7.3767 & 7.8285 & 6.8054 & 7.5743 & 21.2009 & 26.2531 \\
    GPT2 + noise & 7.1437 & 7.9532 & 6.6130 & 7.6918 & 20.9247 & 26.7627 \\[0.5ex]
    \cline{1-7}
    T5  & 12.7026 & 8.6432 & 11.9727 & 8.3622 & 33.8381 & 28.1687 \\
    \textbf{T5 + noise}& \textbf{17.3572} & \textbf{9.8736} & \textbf{16.4354} & 
    \textbf{9.5083} & \textbf{44.0492} & \textbf{30.1061} \\

    \bottomrule
  \end{tabular}
  \vspace{-0.3cm}
\end{table*}
\section{Experiments} \label{exper}

\subsection{Evaluation Metrics} 

Following previous evaluation metrics for slogan generation, we report F1 scores of ROUGE-1 and ROUGE-L  \cite{lin2004rouge}. However, as it is a popular metric for summarization tasks, Rouge score raises concerns that it may not be suitable to assess the quality of a slogan-like generation; distinctiveness, fluency and coherence. Hence, we additionally leverage the Cosine Similarity score based on weights from SimCSE\cite{gao2021simcse}. We adopt Cosine Similarity to measure the similarity between generated slogan with description on how well the to company's identity is incorporated and with reference slogan as an indicator of how much it takes after slogan structure. Here, we assume the higher the Cosine Similarity score, the higher the output slogans are to have stronger alignment with original slogan or input description in that the vectors are represented closer.
\subsection{Analysis \& Discussion}
Table \ref{mainperf} presents the performance of noise perturbation applied to PLMs and compares the results with the two baselines. The first baseline, reconstruction model \cite{misawa2020distinctive}, demonstrates relatively low scores on every metric, including the Cosine Similarity of description. The results suggest that overall generated slogans do not reflect well on their inputs. However, the model demonstrates Cosine Similarity score over 17 with the reference, indicating its ability to generate structurally sound slogan sentences. For instance, in the qualitative example shown in Table \ref{qualitative}, despite the reconstruction model displaying low coherence with the description, the output appears simple and syntactically well-structured as a slogan. This conveys that the Cosine Similarity better comprehends the generated slogans and compensate the evaluation process of using conventional ROUGE score. 

By utilizing the BART, abstractive summarization model as our second baseline \cite{jin2023towards}, it achieved the second-best performance on all metrics. BART, renowned for its abstractive nature, tends to generate longer sentences and reflect key words from descriptions better than any other models. However, when the input embeddings are disrupted with \textit{noise perturbation}, a slight decrease in scores across all metrics is shown in the row of \textit{BART+noise}. We attribute this outcome to the fact that the first five layers of the encoder remaining frozen limits the model's ability to optimize its representations. Although BART generates slogans which lead to high performance scores, there is a lack of consistency in the appropriateness as slogans. For example, the slogans generated by BART in Table \ref{qualitative}, contain phrases like \textit{"The energy of Powerade", "The joy of air Freshener"}. Such phrases correlates with the descriptions about the brand and the firm(i.e., Powerade, Glade), leading to the relatively high scores. However, phrases such as, \textit{"Stay calm and enjoy the ride, love the ending"} are irrelevant which leads to lower scores compared to having only suitable words in those of \textit{T5+noise}.

ROUGE scores for GPT2 exhibit a slight improvement compared to the baselines, while there is a significant increase approximately 13 and 9 scores in each Cosine Similarity. Applying noise perturbation to GPT2 yields minor enhancements in all scores with reference slogans, but reverse phenomenon with description across all metrics. This suggests that perturbing the decoder-only model has minimal impact, as the embeddings are not learned through the encoder layers. Nevertheless, from examples of generated slogans by GPT2 listed in Table \ref{qualitative} (e.g.\textit{The power of love", "Breathing life into your car"}), we observe implicit and metaphoric phrases that could explain the lower Cosine Similarity score in description compared to reference.

Finally, we find a significant performance improvement from the application of the pre-trained T5 model, except for ROUGE scores with reference. Furthermore, the T5 model with noise surpasses all other models with all metrics, exhibiting that replacing input embeddings with Gaussian noise in continuous space not only improves the quality of generated slogans, but also increases the correlation with inputs. In addition, we find that qualitatively in Table \ref{qualitative}, the outputs from T5+Noise comprehends the description better than T5, extracting keywords such as \textit{"power", "sports"} for Powerade; \textit{"fresh"} for Glade. When compared to the original T5 model, the addition of noise generates slogans that exhibit higher coherence with respective descriptions, as evidenced by an increase of over 9 scores in Cosine Similarity. The slogans tend to possess distinct features that align with the expected characteristics from a slogan, as indicated by a 2-point improvement in Cosine Similarity scores when compared to the reference slogans. 

\begin{table*}
  \centering
    \caption{Qualitative results from test dataset}
    \vspace{-0.3cm}
  \begin{tabular}{l|l}
    \hline
    \multicolumn{2}{l}{\textbf{Description:} \label{sample1} \textit{Powerade is a sports drink manufactured and marketed by The Coca-Cola Company...} \hspace{0.5cm}\textbf{Slogan:}\textit{Get up. Stay up.}}\\
    \hline 
    Reconstruction & The power you want to be discovered\\
    BART & The pause that refreshes. The energy ofPowerade. The two are vying for the title:  Stay calm and enjoy the ride, love\\
    &  the ending.Keep calm and wear white crocs when you're hot, hot, comfortable.\\
    GPT2+Noise & The power of love.\\
    T5 & Powerade.\\
    T5+Noise & Powerade is the power of sports.\\ 
    \hline
    \hline
    \multicolumn{2}{l}{\textbf{Description:} \textit{Glade is an S. C. Johnson \& Son brand of household air fresheners that...} \hspace{0.3cm} \textbf{Slogan:}\textit{The joy of fragrance, fresh from Glade.}}\\
    \hline
    Reconstruction & Be legacy.\\
    BART & you can’t even feel them. The joy of air Freshener, even though they’ve been there since 1956. The original Glade\\
    &  fragrance, part of a wholehearted affair, can \\
    GPT2+Noise & Breathing life into your car.\\
    T5 & Reassuringly soft.\\
    T5+Noise & Freshen your life.\\ 
    \hline
  \end{tabular}
  \label{qualitative}
  \vspace{-0.3cm}
\end{table*}

\vspace{-1.5em}
\subsection{Human Evaluation} \label{Human eval text}

To accurately assess the quality of generated slogan, we also conduct human evaluation following the annotation guidelines outlined in \cite{misawa2020distinctive, jin2023towards}, considering three criteria: distinctiveness, coherence, and fluency. We expand the concept of distinctiveness consisting memorable and likeable slogan features to assesses the slogans' catchiness and creativity. Coherence measures how well a slogan aligns with the given description, and fluency evaluates the naturalness and grammatical correctness of the generated slogans, ensuring they make sense. 120 firm/brand description and slogan pairs were evaluated, with a minimum of 10 workers to select a better one between two slogans. For example, all of our questions start with the same prompt: "Which slogan appears to be more..." followed by the distinct, cohered to description, fluent. We show the results of our main model \textbf{T5+0.75\%Noise} compared with two baseline models, GPT2, GPT2+Noise, and T5 without noise in Table \ref{humaneval}. \\

\vspace{-1.5em}
\begin{table}[ht]
  \caption{Human evaluation results. Higher percentage indicates human evaluators perceive \textit{T5+Noise} better over compared models across 120 samples.}
  \vspace{-0.3cm}
  \label{humaneval}
  \begin{tabular}{w{c}{1.9cm}|w{c}{1.8cm}w{c}{1.7cm}w{c}{1.7cm}}
    \toprule
    Models&Distinctiveness&Coherence&Fluency\\
    \midrule
     Reconstruction& 64.03823 & 73.85871 &66.48746\\
     BART & 60.32513 & 68.91377 & 62.34899\\
     GPT2+Noise & 57.43216&73.20675 &64.52005\\
     T5  & 60.08345 & 69.45637 & 64.35574\\ [0.5ex]
     \hline
     Average & 60.46974 & 71.35890 & 64.42806\\
  \bottomrule
\end{tabular}
\vspace{-1.5em}
\end{table}

Our method demonstrates comparable performance in the aspect of coherence with corresponding descriptions, while offering more distinctive results. While previous baselines, reconstruction and BART, aimed to improve the distinctive features, ours outperformed in distinctiveness suggesting our approach generated slogans that contained more creative and memorable phrases.
The scores also support our idea that adding noise to the encoder-decoder structure model is more effective than the decoder-only model by comparing T5+noise and GPT2+noise. Fluency, in general, remains between 62\% $\sim$ 66\%, indicating that our model consistently outperformed others. Noise perturbation to original T5 model proves its effectiveness in enhancing these three criteria, as human evaluators were 60\%, 69\%, and 64\% more likely to choose \textit{T5+Noise} over model without noise. Overall, the evaluators are likely to judge the T5 model with noise perturbation generates slogans that correlate better with company descriptions, achieving an average of score of 71\%, the highest among three measures. To recap, perturbing the input embeddings helped encoder-decoder T5 model to generate distinct, cohered, and fluent slogans compared to reconstruction(non-PLM), BART(summarization model), and GPT2(decoder only model).

\vspace{-0.3em}
\subsection{Ablation Study} \label{absection}
\begin{table}[ht]
  \caption{Ablations of various Gaussian noise perturbation on Cosine Similarity performance.}
  \vspace{-0.1cm}
  \label{ablation}
  \begin{tabular}{p{2.5cm}|w{c}{1.5cm}|w{c}{1.5cm}w{c}{1.5cm}}\toprule
    \toprule
    Level of Noise & Desc & Ref\\
    \midrule
    Noise 0\% & 33.8381 & 28.16867 \\
    Noise 25\% & 41.4585 & \textbf{30.78287}\\
    Noise 50\% & 43.49369 & 29.94495\\
    \textbf{Noise 75}\% & \textbf{44.04924} & 30.10609\\
  \bottomrule
\end{tabular}
\vspace{-0.3cm}
\end{table}

We conduct an ablation study to investigate the impact of varying levels of noise perturbation applied to input embeddings. We consider 25\%, 50\%, and 75\% as our primary choices for the amount of noise, which we solely apply during training. Table \ref{ablation} shows the comparison of Cosine Similarity scores for description and reference of the generated slogans. The original T5 model with 0\% noise perturbation achieves the lowest on both cases, indicating effectiveness of disrupting the input embeddings with Gaussian noise. Moreover, increasing the amount of noise gradually improves the scores for description. Interestingly, for the reference, the highest score is achieved in 25\% noise, with a difference no more than 0.83\% compared to the lowest score by 50\% noise. Ultimately, considering the minor differences in reference score, we select the noise level of \textbf{75\%} as our primary model, which shows the highest performance in description.
\section{Conclusion}

In this work, we introduce a novel slogan generation approach which applies Gaussian noise perturbation to continuous input vectors on pre-trained encoder-decoder model. This model is trained with newly refined slogan-firm/brand level description paired dataset. Extensive experimental results proves that our approach outperforms baseline models and other benchmark models in both quantitative metrics and qualitative evaluations.

\section*{Acknowledgements}
This work was supported by Institute of Information \& communications Technology Planning \& Evaluation (IITP) grant funded by the Korea government (MSIT) (No.2022-0-00908-001, Artificial Intelligence-based Technology for Generating and Recommending Brand or Corporate Identity for Marketing Solutions) and the 2023 Research Fund (1.230036.01) of UNIST (Ulsan National Institute of Science \& Technology).

\clearpage
\bibliography{reference}
\bibliographystyle{ACM-Reference-Format}

\end{document}